\documentclass[a4paper]{article}
\usepackage{Odyssey2022}
\usepackage{graphicx,epsfig,amssymb,amsmath}
\usepackage{tikz}
\usepackage{multirow}
\usepackage{multicol}
\usepackage{hyperref}

\usepackage{subfig}

\usepackage{xcolor}

\def\checkmark{\tikz\fill[scale=0.4](0,.35) -- (.25,0) -- (1,.7) -- (.25,.15) -- cycle;}

\ninept

\title{Impact of Channel Variation on One-Class Learning for Speech Spoof Detection}

\twoauthors{Rohit Arora}
                            {IIIT, Delhi \\
                            \small \tt rohit18115@iiitd.ac.in}
                            {Anmol Arora, Rohit Singh Rathore}
                            {Bennett University, Uttar Pradesh \\
                            \small \tt e19cse010@bennet.edu.in,e19cse040@bennet.edu.in}

\begin{document}
\maketitle

\begin{abstract}
Margin-based losses, especially one-class classification loss, have improved the generalization capabilities of countermeasure systems (CMs), but their reliability is not tested with spoofing attacks degraded with channel variation. Our experiments aim to tackle this in two ways: first, by investigating the impact of various codec simulations and their corresponding parameters, namely bit-rate, discontinuous transmission (DTX) and loss, on the performance of the one-class classification based CM system; second, testing the efficacy of the various settings of margin-based losses for training and evaluating our CM system on codec simulated data.
Multi-conditional training (MCT) along with various data-feeding and custom mini-batching strategies were also explored to handle the added variability in the new data setting and to find an optimal setting to carry out the above experiments. Our experimental results reveal that a strict restrain over the embedding space degrades the performance of the one class classification model. MCT relatively improves performance by 35.55\%, and custom mini-batching captures more generalized features for the new data setting. Whereas varying the codec parameters made a significant impact on the performance of the countermeasure system.

\end{abstract}

\section{Introduction}

Spoofing detection in Automatic Speaker Verification(ASV) systems is a well-established problem as evident from numerous research and competitions held to tackle it \cite{Todisco2019, Kinnunen2017, Zhizheng2015}. Spoofing attacks which are synthetically generated fall in the logical access (LA) category \cite{Yamagishi19-ASV}. Perennially, numerous CMs have been trained and evaluated to give promising results on datasets consisting of speech samples synthesized under controlled and constrained conditions. However, this does not ensure a CM that is robust enough to detect various other spoofing attacks possible in real world scenarios. 

The spoofing attacks encountered in the ASVspoof competitions, namely, in 2015, and 2019, only had non-proactive spoofing attacks. There are many other kinds of spoofing attacks and conditions possible that degrades the performance of the CM system and hence potentially be a threat to the security of the ASV system. \cite{chen2020real, das2020attackers} describes adversarial proactive attacks that were deliberately fabricated to fool the CM system and attack the weak points of the ASV system. \cite{zhang2021initial} talks about partially spoofed audio and showed that CM systems trained for detecting fully spoofed utterances undergo a major degradation in performance when encountered with partially spoofed utterances. The ASVspoof 2021 Logical access database includes bonafide and spoofed samples transmitted over telephony systems which introduces channel variability in the samples \cite{yamagishi2021asvspoof}. The channel variation in the samples makes it much more closer to the actual logical access application scenario.
 
Our work focuses on assessing the impact of channel variation on the performance of CM systems. The work in \cite{Delgado2017} shows the impact of channel variation on LFCC-GMM CM system, but disregards the generalization aspect of it. Apart from that, MCT has proven to be an effective technique to increase the robustness of the model against spoofing attacks degraded or corrupted by various channel or conditional variability \cite{Yu2016, Tian2016, tian2016spoofing, Hanili2016}. The above mentioned works have mostly experimented with environmental noise simulated data, but not much work has been done in case of channel variations.  We think it would be interesting to test whether the CM model can learn features that are invariant to the channel variations when trained on degraded speech samples. 


Many high-performing CM systems, for the ASVspoof 2019 dataset, used various margin-based losses and observed significant performance improvements \cite{lavrentyeva2019stc, Zhang_2021, zhang2021empirical, wang2021comparative}. These approaches have shown promising results.  However, with the data being degraded by codec simulations, the margin values used may create a restrain on the embedding space that could lead to the model overfitting on spoofed and bonafide samples and might not be optimal for the task at hand.  We think it would be interesting to compare the efficacy of \emph{Softmax}, \emph{AM-Softmax}, \emph{OC-Softmax}, and various other settings of these losses on this new and relatively more practical codec simulated dataset. 


Previously experiments have been carried out by varying the frame size and hop length at the time of feature extraction \cite{AhmedSocialSense} , but as per our knowledge negligible amount of empirical analysis is done by varying the length of the speech sample and custom mini-batching strategies in the domain of speech spoof detection. This leads us to consider the following questions: \emph{'What is the optimal data-feeding and mini-batching strategy for handling the added variability introduced due to external conditions?'} To investigate this question, we used random and various custom batching strategies to train our model on two datasets with different level of variability, both derived from ASVspoof 2019 dataset \cite{Yamagishi19-ASV, Delgado21-ASV}. 

In this paper, we partly extend on previously mentioned \cite{Delgado2017} work by carrying out the experiments using one-class classification Resnet model on more relevant ASVspoof 2019 train set and also cover the generalizability aspect as it is an accurate measure of how well the model will perform in real world scenarios. We did that by testing it on the recently released ASVspoof 2021 evaluation set. Various data-feeding and mini-batching strategies were explored to handle the added variability in the new data setting. The optimal strategy is then used to investigate the impact of various codec simulations and their corresponding parameters. And finally we test the efficacy of the various settings margin based losses for training and testing on codec simulated data.

The organisation of the paper is as follows. Section \ref{loss function based studies} discusses about the margin based losses used for training the CM system. Section \ref{Codec} gives a brief explanation about the codecs used to simulate the data used for experimentation. Section \ref{experimental setup} talks about the datasets, data-feeding and custom mini-batching strategies used for training the CM system. Section \ref{Results and Discussion} states the result and inferences that can be made from them. Finally, section \ref{Conclusion} and \ref{Acknowledgements} concludes the paper and states the acknowledgments respectively.



\section{Loss function based studies}\label{loss function based studies}

Margin-based Softmax loss has gained popularity in the field of spoof detection, and they have the ability to enhance feature discrimination by increasing the feature margin between different classes. Their unique upside being increased optimization stability and clear geometric interpretability. In this section, we first briefly go through Softmax loss (\ref{Softmax loss}) and the losses commonly used in the field of spoof detection (\ref{ams and ocs}). 

\subsection{Softmax loss} \label{Softmax loss}

Here is the formulation of Softmax loss. 
\begin{equation}
\mathcal{L}_{S}=\frac{1}{N} \sum_{i=1}^{N} \log \left(1+e^{\left(\boldsymbol{w}_{1-y_{i}}-\boldsymbol{w}_{y_{i}}\right)^{T} \boldsymbol{x}_{i}}\right),
\end{equation}
where $N$ is the number of samples in a mini-batch, $\boldsymbol{x}_{i}$ $\in$ ${R}^{D}$ and $y_{i}$ $\in$ $\{0,1\}$ are the embedding and labels respectively. $\boldsymbol{w}_{0}$,$\boldsymbol{w}_{1}$ $\in$ ${R}^{D}$ are weight vectors for the two classes. 0 and 1 pertaining to bonafide and spoofed class respectively. 

\subsection{Margin based Softmax loss} \label{ams and ocs}
The Equation (\ref{ams})and (\ref{ocs}) formulates the AM-Softmax and OC-Softmax loss respectively.

\begin{equation} \label{ams}
\mathcal{L}_{\textit{AMS}}=\frac{1}{N} \sum_{i=1}^{N} \log \left(1+e^{\alpha\left(m-\left(\hat{\boldsymbol{w}}_{y_{i}}-\hat{\boldsymbol{w}}_{1-y_{i}}\right)^{T} \hat{\boldsymbol{x}}_{i}\right)}\right)
\end{equation}

\begin{equation} \label{ocs}
\mathcal{L}_{\textit{OCS}}=\frac{1}{N} \sum_{i=1}^{N} \log \left(1+e^{\alpha\left(m_{y_{i}}-\hat{w}_{0} \hat{x}_{i}\right)(-1)^{y_{i}}}\right)
\end{equation}

In the equations (\ref{ams}) and (\ref{ocs}) , $\hat{x_{i}}$ stands for normalized input vector containing Linear Frequency Cepstral Coefficients (LFCC) speech features and $y_{i}$ stands for the output labels of the ${i}$-th sample. The $\hat{w_{0}}$ is the normalised weight vector which optimizes direction of the target class embedding. This loss function uses two margins $(m_{0},m_{1}\in[-1,1],m_{0}>m_{1})$ to bound the compact space for the target class in the direction of $\hat{w_{o}}$ and have a wider angular margin for non-target class. Further details can be found in~\cite{Zhang_2021}.

These losses improve upon the Softmax loss by introducing a margin. The AM-Softmax makes the embedding distribution compact for both classes. At the same time, OC-Softmax compacts the embedding space only for the bonafide class. This strategy avoids over-fitting on known spoofed classes and makes the latter more suitable for the task of spoof detection. However, due to channel variation, additive noise might increases the angular domain of all samples, even genuine speech, in the embedding space. This makes it essential to compare the performance of basic Softmax loss with OC-Softmax. As there are chances, the smaller angles for embedding space of genuine speech might lead to a decrease in performance. Thus, we will test the Softmax loss and variations of angle with OC-Softmax loss functions in this paper.

\setlength{\tabcolsep}{3pt}
\renewcommand{\arraystretch}{1}
\begin{table}[h!]
\caption{List of codecs and their corresponding parameters, used to degrade the dataset, categorized according to their usage. The checkmark in the corresponding cell indicates the availability of the parameter for tuning}
\label{table:1}
\begin{center}
\begin{tabular}{ |c|c|c|c|c|c| } 
\hline
\multirow{2}{4em}{Usage} & \multirow{2}{4em}{Codecs} & \multicolumn{4}{c|}{Parameters} \\
\cline{3-6}
& & Bit-rate & Loss & mu/a-law & DTX\\
\hline
\multirow{2}{4em}{Landline} & G.711 & & & \checkmark &\\ 
\cline{2-6}
& G.726 & \checkmark & & \checkmark & \\ 

\hline
\multirow{3}{4em}{Cellular} & AMR-NB & \checkmark & & & \checkmark\\ 
\cline{2-6}
& AMR-WB  & \checkmark & & & \checkmark \\ 
\cline{2-6}
& GSM-FR & & & &\\ 
\hline
\multirow{3}{4em}{Satellite} & G.728 & & \checkmark & & \\ 
\cline{2-6}
&  CVSD & \checkmark & & &\\ 
\cline{2-6}
& Codec2 & \checkmark & & &\\ 
\hline
\multirow{4}{4em}{VoIP} & SILK & \checkmark & \checkmark & & \checkmark\\ 
\cline{2-6}
& SILK-WB & \checkmark & \checkmark & & \checkmark\\ 
\cline{2-6}
& G.729a & & & &\\ 
\cline{2-6}
& G.722 & \checkmark & & &\\
\hline
\end{tabular}
\end{center}
\end{table}

\section{Codec Simulations}\label{Codec}

To tackle this newly added complexity systematically, we need to be aware of the types of codec simulations and their corresponding parameters that can affect the performance of the countermeasure system. For both the data-sets the additional channel variation were simulated with the Idiap acoustic simulator software which is available online at \href{https://github.com/idiap/acoustic-simulator}{$<$https://github.com/idiap/acoustic-simulator$>$}. Here's a brief description of the parameters available for codec simulation:

\begin{itemize}
    \item \textbf{Bit-rate} : Bit-rate alludes to the quantity of bits that are passed on or processed in a given unit of time. The higher the bit-rate the more detailed the audio and hence better the audio quality of the output. 
    \item \textbf{Loss} : The loss parameter gives control over the amount of packets lost during the transmission. 
    \item \textbf{mu/a-law} : Common companding algorithms used in telephony system.  Both have fairly minimal difference and the advantages of one over the other are insignificant, with $\mu-law$ having higher dynamic range but also has higher distortion for small signals when compared to $a-law$.
    \item \textbf{DTX} : DTX stands for discontinuous transmission, it diminishes the transmission rate during inactive discourse periods while maintaining a respectable level of yield quality.
\end{itemize}

We hypothesize that the above discussed parameters decides how informative the audio output is and have a direct relation to the performance of the CM system for spoof detection. These parameters are not common to all the codec simulations, hence we have listed the codecs along with their parameters available for tuning, categorized according to usage, used to create the modified dataset in table \ref{table:1}

\section{Experimental Setup} \label{experimental setup}

This section describes the model implementation details and the different data-feeding strategies and custom mini-batching strategies used for training the model. The experiments were carried out on two degraded datasets that were derived from ASVspoof 2019 dataset. All the experiments in this section are done with Resnet-OC, as it showed comparable performance with state-of-the-art ensemble systems \cite{chettri2019ensemble, yang19b_interspeech} being a single system.

\subsection{Implementation details}

The model architecture is adapted straight away from the one-class classification model(Resnet-OC). All the parameters were kept the same, with the Adam optimizer being used to update the weights. For pre-processing, we extract 60-dimensional LFCCs (including delta and double deltas) from the audio samples. The frame size was set to approximately 20 ms, and the hop size was 10 ms (50\% overlap). Pytorch based LFCC layer was embedded into the original model. The GitHub link of the original OC model and our implementation are \href {https://github.com/yzyouzhang/AIR-ASVspoof}{$<$https://github.com/yzyouzhang/AIR-ASVspoof$>$} and \href{https://github.com/rohit18115/ASVspoof2021_OC_model}{$<$https://github.com/rohit18115/ASVspoof2021\_OC\_model$>$} respectively.

\subsubsection{Dataset} \label{Dataset}

\renewcommand{\arraystretch}{1}
\setlength{\tabcolsep}{6pt}
\begin{table*}[t]
\caption {Logical access results of LFCC-Resnet with OC-Softmax loss trained on 3 versions of datasets(\textit{Original}, \textit{Ver-1} and \textit{Ver-2}) tested on our degraded development(\textit{deg-dev}) set, degraded ASVspoof 2019(\textit{sim19})  and organisers evaluation set(\textit{eval21}).} 
\label{table:big}
\begin{tabular}{lllllllllllll}
\hline
\hline
\multirow{3}{4em}{Dataset} & \multirow{3}{4em}{Mini-Batch} & \multicolumn{11}{c}{Data-feeding Strategies}\\
\cline{3-13}
& & \multicolumn{3}{c}{$1 sec$} &  & \multicolumn{3}{c}{$mean$} & & \multicolumn{3}{c}{$max$}    \\
\cline{3-5}\cline{7-9}\cline{11-13}
&   &\textit{deg-dev}& \textit{sim19} & \textit{eval21} & & \textit{deg-dev} & \textit{sim19}  &  \textit{eval21} & & \textit{deg-dev} & \textit{sim19} &   \textit{eval21}  \\ 
\hline
\hline

\multirow{3}{4em}{\textit{Original}} & Random & 24.12&	26.65&	33.36&		&11.04&	12.16&	34.38&		&3.27&	4.34&	38.47\\
\cline{3-13}
 & Custom class & 25.84&	26.52&	25.91&		&13.50&	14.20&	32.54&		&4.30&	4.78&	34.56 \\
  & Custom speak & 24.50&	27.14&	25.80&		&11.98&	12.18&	32.14&		&4.88&	5.06&	35.70 \\

\hline
\multirow{2}{4em}{\textit{Ver-1}} & \textbf{Random} & \textbf{18.58}	& \textbf{20.32}	& \textbf{22.74} &		&4.28	&7.11	&30.90&		&0.61	&0.88	&37.56 \\
\cline{3-13}

 & Custom sim & 23.52	&24.02	&24.40&		&8.02	&23.03	&26.20&		&2.55	&27.56	&28.78 \\
\hline
\multirow{2}{4em}{\textit{Ver-2}} & \textbf{Random} & \textbf{19.11}	& \textbf{21.70}	& \textbf{23.86}&		&5.76	&9.54	&31.35&		&0.79	&1.08 & 38.35\\
\cline{3-13}
 & Custom sim & 24.28	&24.95	&24.66&		&8.91	&16.44	&26.91&		&3.69	&20.65	&27.21 \\
\hline
\multicolumn{2}{l}{\textbf{Fusion}} & \textbf{18.05} & \textbf{19.68} & \textbf{21.50} \\

\hline
\hline
 
\end{tabular}
\end{table*}

The ASVspoof 2019 LA, ASVspoof 2019 LA-Degraded and ASVspoof 2021 LA dataset are used for the experimentation. Where the LA-Degraded dataset is our codec simulated version of the original 2019 LA dataset. Brief description for the dataset are given as follows: 

\textbf{ASVspoof 2019 LA} : The LA track for ASVspoof 2019 contains bonafide, and spoofed speech data generated using 17 different text-to-speech (TTS) and voice conversion (VC) systems. Six of these systems are designated as known attacks,  with the other 11 being designated as unknown attacks. For more details about the dataset and rules, refer to \cite{wang2020asvspoof} \cite{Delgado21-ASV} 

\indent\textbf{ASVspoof 2019 LA-Degraded} : The train and development set of the ASVspoof 2019 LA dataset is joined, and randomly \emph{10,000} samples are chosen for the development set, and the rest \emph{40,224} samples are used as a train set. The models are trained on two datasets. Both the train sets are constructed to test the performance of our model under different settings of bit-rate, Loss, DTX, and $\mu/a-law$ parameters for various codec simulations. A list of 16 and 45 codec simulations was applied on the original dataset in a cyclic manner for the first and second datasets, respectively. This implies that there are \emph{2,514} and \emph{893} samples per degradation for the first and second dataset, respectively. This setup provides the chance to test the effect of random and custom batching to handle different levels of variability introduced in the two datasets through codec simulations.

\textbf{ASVspoof 2021 LA} \cite{yamagishi2021asvspoof} : The dataset released for ASVspoof 2021 competition only consisted of the evaluation set which entails new bonafide and spoofed trials of the same speakers in the VCTK corpora which is available online at \href{https://doi.org/10.7488/ds/1994}{$<$https://doi.org/10.7488/ds/1994$>$}. The spoofed utterance were synthesized using the same attacks in the 2019 evaluation set. All the utterance were transmitted over telephony systems including VoIP, PSTN and various others. The fact that the presence of new trial utterances, and unknown spoofing attacks and codec degradation's make it a good choice for using it as our testing set to evaluate the generalisation capabilities of our model.

The randomly sampled development set and ASVspoof 2019 LA eval set are also degraded in a cyclic manner and are used for validating our model. Whereas, \emph{ASVspoof 2021 eval} set is used as test set. \textit{Ver-1}, \textit{Ver-2}, \textit{deg-dev}, \textit{sim19} and \textit{eval21} are names for first, second, simulated development and evaluation set respectively.

\subsubsection{Custom mini-batching strategies} \label{Custom mini-batching strategies}

Moreover, random and custom mini-batching strategies were used to investigate the effect of different levels of variability introduced in the two datasets. The three types of custom mini-batching used are briefly described as follows:

\begin{itemize}
    \item \textbf{Custom class}:  Every mini-batch has an equal number of spoofed and bonafide samples.
    \item \textbf{Custom speak}: Every mini-batch has an equal number of spoofed and bonafide samples, and that for every spoofed sample, there is a bonafide sample of the same speaker.
    \item \textbf{Custom sim}: Every mini-batch has an equal number of spoofed and bonafide samples, and that for every spoofed sample, there is a bonafide sample of the same codec simulation.
\end{itemize}


 This way, we make sure that the model is only learning to distinguish between spoofed and bonafide samples and is not biased towards the majority class. It also ensures that the model's learning is not confused by the characteristics of various speaker or codec simulations depending on the custom mini-batching.

\subsubsection{Data-feeding strategies} \label{Data-feeding strategies}
 In order to give a fair analysis we cover the whole spectrum of length by taking into account the minimum, mean and maximum sample length for each batch used in training. A brief description is as follows:

\textbf{One-second chunks} The speech samples are sliced into 1-second chunks(16000 samples) and then randomly selected and fed to the model.

\textbf{Max sample length} The speech samples are randomly selected to form a batch, and all the speech samples are repeat padded to equal the length of the sample with maximum length.

\textbf{Mean sample length} The speech samples are randomly selected to form a batch, and all the speech samples are sliced/repeat padded to equal the mean length of the samples in the batch.

\section{Results and Discussion}\label{Results and Discussion}


\begin{figure*}
\begin{tabular}{cc}
\subfloat[Impact of different categories of codec simulations\label{categories}]{\includegraphics[width = 3.1in]{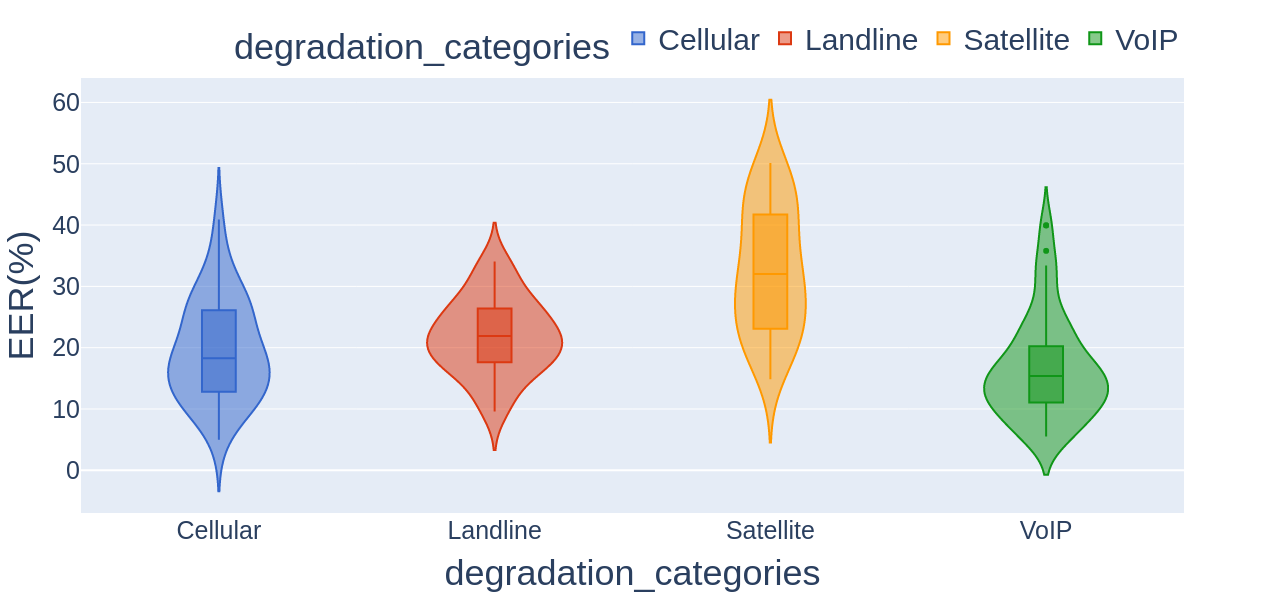}} &
\subfloat[Performance comparison of the systems mentioned in Table \ref{table:big}\label{systems}]{\includegraphics[width = 3.1in]{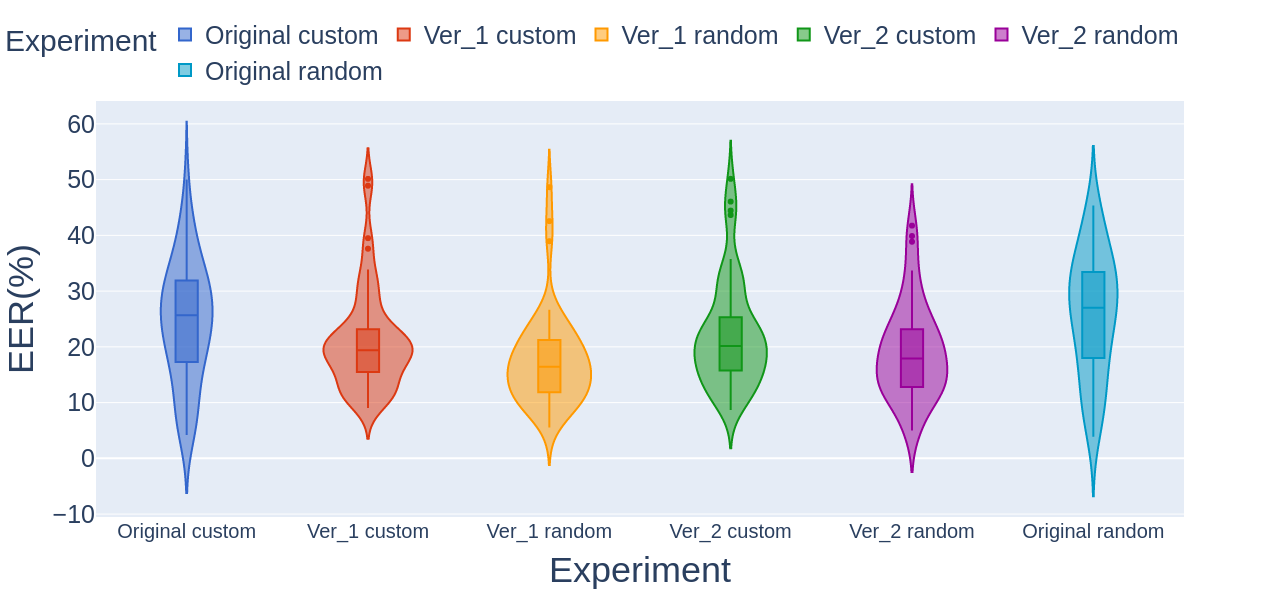}} \\
\subfloat[Effect of bit-rate and discontinuous transmission(DTX)\label{bitrate}]{\includegraphics[width = 3.1in]{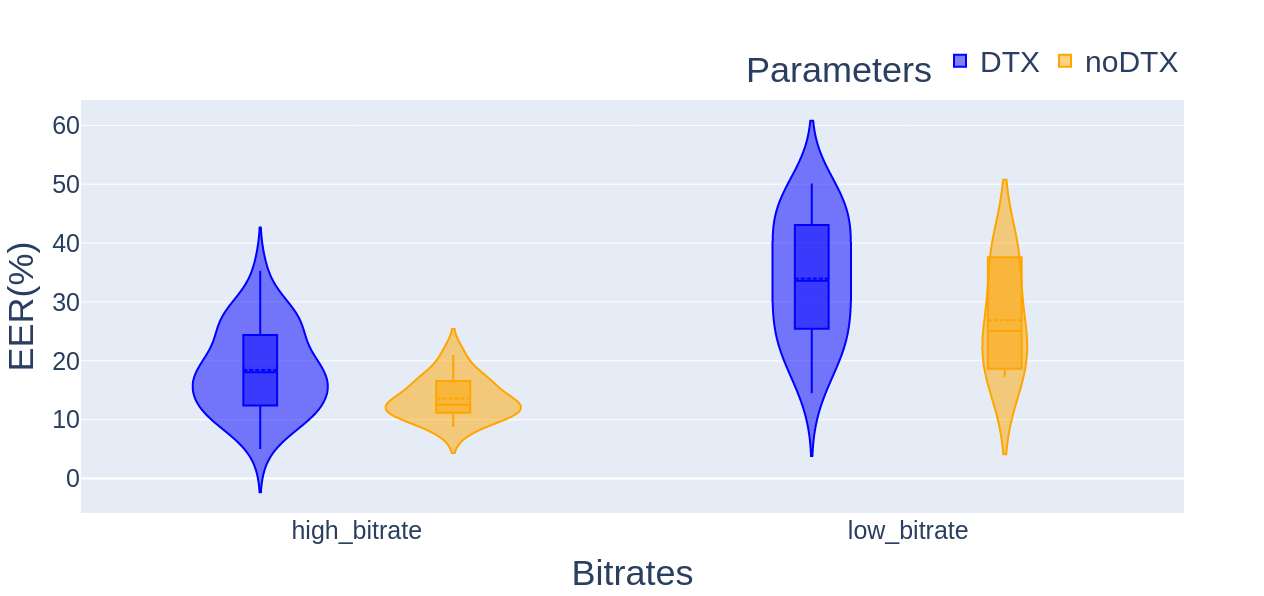}} &
\subfloat[Effect of bandwidth and packet loss\label{bandwidth}]{\includegraphics[width = 3.1in]{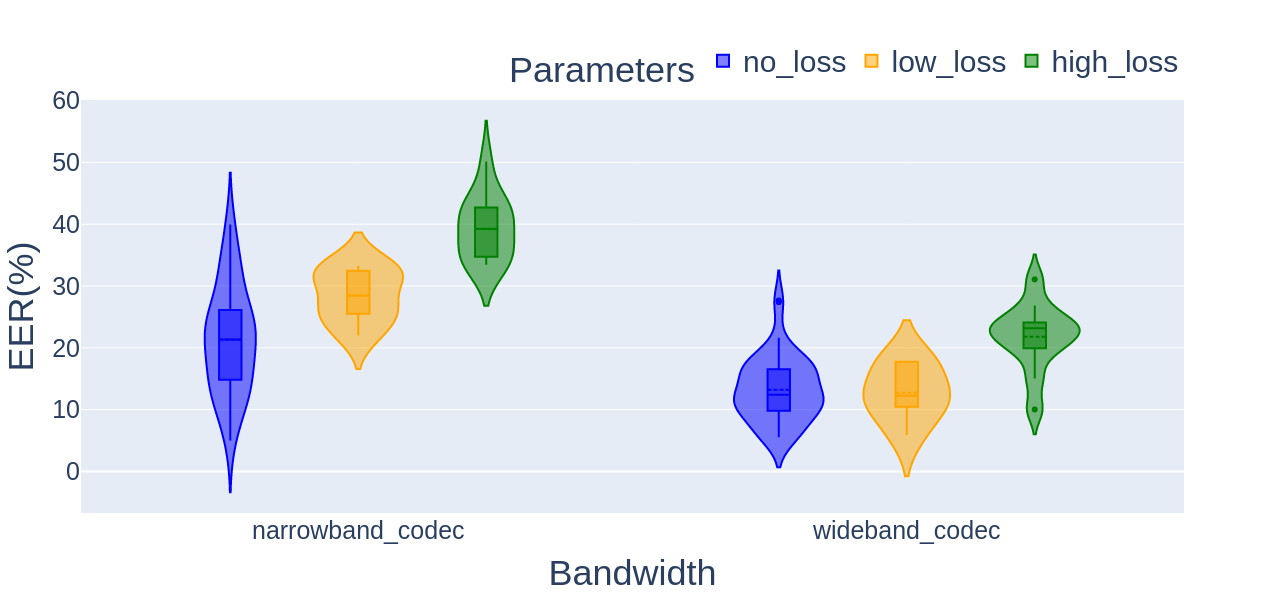}}\\
\end{tabular}
\caption{Violin plots to show the impact of channel variation on one-class classification model.}
\label{image:violinplot}
\end{figure*}

\setlength{\tabcolsep}{5pt}
\begin{table}[h!]
\caption {Logical access results of One-Class Learning trained on original ASVspoof 2019 LA dataset and tested on our degraded and original datasets.} 
\label{table:small}
\begin{tabular}{lllllll}
\hline
\hline
 length & Mini-Batch & \multicolumn{2}{c}{$Degraded$} &  & \multicolumn{2}{c}{$Original$} \\ 
\cline{3-4} \cline{6-7} 
 &  &  Dev & Eval  & &  Dev & Eval\\    
\hline
\multirow{2}{2em}{$1sec$} &Random & 24.12  & 26.65  &  & 11.62 & 13.54\\

                            & Custom class & 25.84 & 26.52  & & 12.40 & 13.31 \\
                            & Custom speak & 24.50 & 27.14 & & 12.28 & 12.45 \\
\cline{2-7} 
\multirow{2}{2em}{$mean$} &Random & 11.04  & 12.16  &  & 7.24 & 10.30\\

                            & Custom class & 13.50 & 14.20 & & 7.90 & 9.25 \\
                            & Custom speak & 11.98 & 12.18 & & 8.02 & 9.33 \\
\cline{2-7} 
\multirow{2}{2em}{$max$} &\textbf{Random} & \textbf{3.37}  & \textbf{4.34}  &  & \textbf{0.46} & \textbf{1.23}\\

                            & Custom class & 4.30 & 4.78 & &1.03 & 1.49   \\
                            & Custom speak & 4.88 & 5.06 & & 1.27 & 1.50 \\

\hline
\hline
 
\end{tabular}
\end{table}


All the results are presented in terms of Equal Error Rate (EER[\%]), and the minimum tandem detection cost function (min-tDCF) is omitted. This section can be further divided into two parts. First, section \ref{Preliminary experiments}, which talks about the impact of multi-conditional training and the effect of various data-feeding and mini-batching strategies mentioned in section \ref{Data-feeding strategies} and \ref{Custom mini-batching strategies}. These are portrayed through the results logged in Table \ref{table:big} to \ref{table:small}. The work in this section helps us select the optimal setting for the further experiments and leads us to the second section \ref{Impact of channel variation}. It discusses about the impact of various channel variations and its corresponding parameters mentioned in section \ref{Codec} on different loss functions and their settings.  the The Figure \ref{image:violinplot} illustrates the performance of the systems mentioned in the Table \ref{table:big}. Whereas the Table \ref{table:4} gives an analysis of LFCC-Resnet model trained with Softmax and various settings of OC-Softmax loss. 

\subsection{Preliminary experiments} \label{Preliminary experiments}

The result obtained on our degraded ($deg\ dev$ and $sim19$) and original, development set and evaluation sets are presented in the Table \ref{table:small}. It states that the One-Class classification model trained on the Original ASVspoof 2019 LA dataset incurs a significant loss in performance as compared to when tested on original dataset, this is irrespective of the data-feeding strategies used. The results logged in Table \ref{table:big} are obtained by training our model on original, \textit{Ver-1} and \textit{Ver-2} datasets and testing on \textit{deg-dev}, \textit{sim19} and \textit{eval21} datasets. These experiments gave us a chance to test the efficacy of MCT, and since \textit{eval21} is a recently released dataset by the ASVspoof community, its distribution is different from the datasets used from training our model. Hence, it also tests the generalization capabilities of the different data-feeding and mini-batching strategies used to train the model. It is worth mentioning that we have not used the pre-trained weights given by the author to avoid any inconsistencies with other models we have trained. 
 
 From the experiments carried out, it can be confirmed that the length of the speech sample and the mini-batching strategy used does affect how well the model generalizes on different datasets. If we compare the performance of the models with respect to the data-feeding strategies, it is evident that the model loses its generalizing capabilities with an increase in the sample size that is being fed to it. It can be observed that the multi-conditional training with random mini-batching done on \textit{Ver-1} and \textit{Ver-2} does improve performance as compared to when the model is trained on the original dataset. It is also worth mentioning that even though random-batching showed more performance gain as compared to custom batching, the latter helped the model extract features that helped the model generalize well on the evaluation set. Furthermore, this is consistent along with all the data-feeding strategies. Although models trained using $max$ length do not generalize well on new datasets, from table \ref{table:small} it can be observed that they still show remarkable performance on degraded and original, Development (\textit{deg-dev}) and Evaluation set (\textit{sim19}). On the other hand, the model trained on $1 sec$ samples perform better on the \textit{eval21} dataset as compared to the models trained using $mean$ and $max$ data-feeding strategies. 
 
 The above observations seems intuitive since the larger the variability (speaker or degradation based characteristics) in each batch the more difficult it will be for the model to pick up generalized features in case of random mini-batching. Hence \emph{custom sim} seems to capture more generalized features and has relatively small gap in performance when evaluated on development set and evaluation set as compared to \emph{Random}. As far as the observations for different data-feeding strategies are concerned, the results lead us to infer that with increase in sample size the model starts to over-fit and hence show poor performance on any data with different distribution from the data that is fed to it at the time of training.


\subsection{Impact of channel variation} \label{Impact of channel variation}


 Since the model trained on $1 sec$ chunks relatively captures more generalized features its results were used to plot the violin plots in the Fig. \ref{image:violinplot}. The Fig. \ref{categories} describes the impact of different categories of codec simulations. The Fig. \ref{systems} gives a comparison of the performance of the different data-feeding and custom mini-batching experiments. The Fig. \ref{bitrate} shows the impact of the parameters bit-rate and DTX on the performance of the systems. Whereas Fig. \ref{bandwidth} shows the impact of bandwidth and packet loss. The following observations are made:

\begin{itemize}
    \item \textbf{All the systems performed well on wide-band codecs as compared to the narrow-band ones.} This can be attributed to the relative wide range of frequency bandwidth used to represent the audio in case of wide-band codec simulation. To be specific the human voice extends from 80 Hz to 14 kHz, but traditional, narrowband telephone calls limit audio frequencies to the range of 300 Hz to 3.4 kHz. Wideband audio relaxes the bandwidth limitation and transmits in the audio frequency range of 50 Hz to 7 kHz and hence is able to represent both bonafide and genuine speech much more accurately then narrowband codecs.
    
    \item \textbf{The EER increases in proportion to the increase in loss parameter.} The Subjective Quality Measure(SQM) of the audio is directly proportional to packet loss \cite{Laghari2019}, and in case of spoof detection it would decrease the SQM of both bonafide and spoofed samples. And from the results we can infer that the packet loss not only also reduces the SQM but also the amount of valuable artifacts that help distinguish spoofed and genuine samples.
    
    \item \textbf{Codecs with high bit-rate setting makes spoof detection easier as compared to low bit-rate ones.} Bit-depth indicates the number of bits used to create each sample in the audio. Bit-rate is directly proportional to bit-depth. So lower the bit-rate, less detailed the audio. Which in-turn decreases the amount of relevant information needed for spoof detection.  With similar logic if DTX parameter is not used, it should increase the performance of the systems, this is clearly true only codec with high bit-rates but not as much for codec with the low bit-rate setting.  
    
    \item \textbf{Sattelite-based codec simulations showed an exceptionally high EER.} Both \emph{CSVD} and \emph{C2} codecs encodes at very low bits per sample for the audio. This is done to conserve bandwidth over tactical links which are used for various high security tasks. For instance, a \emph{CSVD} sampled at 16 kHz is usually encoded at 16 kbits/s, that means it encodes at 1 bit per sample and as we can infer from our results, its not enough information for spoof detection.  
\end{itemize}  

From Figure \ref{systems} and Table \ref{table:big} it is evident that a weighted score fusion between systems trained on Ver 1 with random batching and Ver 2 with random batching would increase the performance. After few experiments, we found out that the weight of 10 and 90 respectively gave the best performance. 

\setlength{\tabcolsep}{4pt}
\begin{table}[h!]
\caption {Logical access results of One-Class Learning trained on original ASVspoof 2019 dataset and tested on our degraded development set and original development set.} 
\label{table:4}
\begin{tabular}{lllll}
\hline
\hline
 Model & \textit{dev-deg} & \textit{sim19}  & \textit{eval21} \\    
\hline
\hline

Softmax & 20.06 & 22.88 & 26.67 \\
AM-Softmax($\boldsymbol{m}$=0.9) & 25.24 & 27.55 & 28.68 \\
AM-Softmax($\boldsymbol{m}$=0.3) & 19.85 & 21.90 & 23.40 \\
OC-Softmax($\boldsymbol{m_0}$=0.9, $\boldsymbol{m_1}$=0.3) & 19.11 & 21.70 & 23.86 \\
\textbf{OC-Softmax($\boldsymbol{m_0}$=0.5, $\boldsymbol{m_1}$=0.2)} & \textbf{19.46} & \textbf{19.88} & \textbf{22.75} \\

\hline
\hline
 
\end{tabular}
\end{table}

The EER[$\%$] values of the various settings of loss functions are reported in Table \ref{table:4}. Note, model were trained using $1sec$ chunks and random batching for these comparisons. The comparison is made between OC-Softmax-wide($\boldsymbol{m_0}$=0.4, $\boldsymbol{m_1}$=0.2) which is the original version used in \cite{Zhang_2021} and is optimized for ASVspoof2019 dataset, the OC-Softmax-wide($\boldsymbol{m_0}$=0.5, $\boldsymbol{m_1}$=0.2), which is the loss with less restricted embedding space for genuine samples, two settings of AM-Softmax loss and finally Softmax loss. From the EER values it can be inferred  that the strict restrain over the embedding space of genuine samples has adverse effect and reducing the restrain might be a better option for the new data setting. And the equal margin for both the classes for AM-Softmax performs better then Softmax but relatively bad then both the versions of the OC-Softmax loss again shows the utility of One-class classification in the this field. 

\section{Conclusions} \label{Conclusion}

  In this paper, we constructed two databases, adding various codec simulations to facilitate our experiments. Results show that the performance of the one-class classification system suffers in the new setting. Encouragingly, multi-conditional training improves performance by 35.55\% . It is observed that random mini-batching gives lower EER as compared to custom mini-batching, whereas the latter generalizes well then the former for the evaluation set. Moreover, it can be confirmed that the length of the speech sample and the mini-batching strategy used to decide how well the model generalizes on different datasets. And a strict restrain on the embedding space over the genuine samples leads to sub-optimal performance and reducing the restrain would be a good option to deal with the added variability due to codec simulations.     

\section{Acknowledgements} \label{Acknowledgements}

This work would not have been possible without the guidance of Dr. Saket Anand and Dr. Aanchan Mohan. Special thanks to You Zhang, his enthusiasm and attention to detail helped improve the structure and presentation of the paper in short span of time.


\bibliographystyle{IEEEbib}
\bibliography{Odyssey2022_BibEntries}

\end{document}